\documentclass[letterpaper, 10 pt, conference]{ieeeconf}  

\IEEEoverridecommandlockouts                              

\usepackage{graphicx}
\usepackage{wasysym}
\usepackage{booktabs}
\usepackage{balance}
\usepackage{multirow}
\usepackage{overpic}
\usepackage{bbm}
\usepackage{dsfont}
\usepackage{caption}
\usepackage{color}
\usepackage{pifont}

\newcommand{\PAR}[1]{\vskip4pt \noindent {\bf #1~}}  

\newcommand{\ourmethod}{DR-SPAAM}

\newcommand*\widebar[1]{%
   \hbox{%
     \vbox{%
       \hrule height 0.5pt 
       \kern0.3ex
       \hbox{%
         \kern-0.1em
         \ensuremath{#1}%
         \kern-0.1em
       }%
     }%
   }%
}

\title{\LARGE \bf
DR-SPAAM: A Spatial-Attention and Auto-regressive Model for \\Person Detection in 2D Range Data}

\author{Dan Jia$^{1}$, Alexander Hermans$^{1}$, and Bastian Leibe$^{1}$
\thanks{$^{1}$All authors are with the Visual Computing Institute, RWTH Aachen University.
        {\tt\small lastname@vision.rwth-aachen.de}}%
}

\usepackage[dvipsnames]{xcolor}
\usepackage{graphicx}
\usepackage{amssymb}
\usepackage{multirow}
\usepackage{colortbl}
\usepackage{color}
\usepackage{hyperref}
\usepackage{xspace}
\usepackage{tabularx}
\usepackage{tikz}
\usetikzlibrary{calc,shapes,arrows}
\usepackage{pgfplots}
\usepackage{siunitx}
\usepackage{booktabs}
\usepackage{csvsimple}
\usepackage{makecell}

\definecolor{wd_c}{HTML}{E24A33}
\definecolor{wc_c}{HTML}{348ABD}
\definecolor{wa_c}{HTML}{D175F0}
\definecolor{wp_c}{HTML}{8EBA42}
\definecolor{awesome_orange}{HTML}{F9A91F}
\definecolor{good_gray}{HTML}{777777}
\definecolor{deep_pink}{HTML}{FF69B4}
\definecolor{shit_brown}{HTML}{8B4513}
\definecolor{random_blue}{HTML}{7DE2F0}
\definecolor{evil_green}{HTML}{366D32}

\pgfplotsset{
    precrec/.style={
        inner sep=0pt,outer sep=0pt,
        ylabel style={font=\scriptsize,yshift=-18pt},
        xlabel style={font=\scriptsize,yshift=6pt},
        width={1.1\linewidth},
        height={1.1\linewidth},
        yticklabel style = {font=\scriptsize,xshift=-0.3ex},
        xticklabel style = {font=\scriptsize,yshift=-0.3ex},
        legend image post style={line width =1.5pt},
        tick label style={/pgf/number format/assume math mode=true},
        tick align=outside,
        major tick length=2pt,
        every tick/.style={black, thin},
        tick pos=left,
        axis line style={draw=none},
        xtick={0,20,...,100},
        ytick={0,20,...,100},
        xlabel={Recall [\%]},
        ylabel={Precision [\%]},
        xmin=-2,
        xmax=102,
        ymin=-2,
        ymax=102,
        axis background/.style={fill=black!11!white},
        grid=both,
        grid style={white}
    }
}

\pgfplotsset{legend image code/.code={%
                \draw[mark repeat=2,mark phase=2]
                plot coordinates {
                (0cm,0cm)
                (0.3cm,0cm)        
                (0.6cm,0cm)         
                };%
            }
}

\def\addlegendimage{\csname pgfplots@addlegendimage\endcsname}

\newcommand{\sampleselect}[1]{data/#1_fast.csv}
\newcommand{\precrec}[3]{\addplot [color=#2, #3] table [y=prec, x=rec, col sep=comma] {\sampleselect{#1}};}
\newcommand{\precrecnl}[3]{\addplot [color=#2, #3] table [y=prec, x=rec, col sep=comma,forget plot] {\sampleselect{#1}};}

\makeatletter
\DeclareRobustCommand\onedot{\futurelet\@let@token\@onedot}
\def\@onedot{\ifx\@let@token.\else.\null\fi\xspace}

\def\eg{\emph{e.g}\onedot} 
\def\ie{\emph{i.e}\onedot} 
\def\cf{\emph{c.f}\onedot}

\def\etal{\emph{et al}\onedot}
\makeatother

\newcolumntype{Y}{>{\centering\arraybackslash}X}

\begin{document}

\maketitle
\thispagestyle{empty}
\pagestyle{empty}

\begin{abstract}

Detecting persons using a 2D LiDAR is a challenging task due to the low information content of 2D range data.
To alleviate the problem caused by the sparsity of the LiDAR points, current state-of-the-art methods fuse multiple previous scans and perform detection using the combined scans.
The downside of such a backward looking fusion is that all the scans need to be aligned explicitly, and the necessary alignment operation makes the whole pipeline more expensive -- often too expensive for real-world applications.
In this paper, we propose a person detection network which uses an alternative strategy to combine scans obtained at different times.
Our method, Distance Robust SPatial Attention and Auto-regressive Model (\ourmethod), follows a forward looking paradigm.
It keeps the intermediate features from the backbone network as a template and recurrently updates the template when a new scan becomes available.
The updated feature template is in turn used for detecting persons currently in the scene.
On the DROW dataset, our method outperforms the existing state-of-the-art, while being approximately four times faster, running at 87.2~FPS on a laptop with a dedicated GPU and at 22.6~FPS on an NVIDIA Jetson AGX embedded GPU.
We release our code in PyTorch and a ROS node including pre-trained models.
\end{abstract}

\section{INTRODUCTION}
\label{sec:introduction}

Detecting persons in the surrounding environment is a key requirement for many robotic applications including search and rescue, security, and health care.
Currently, this is often accomplished using multiple RGB(-D) cameras, in combination with a deep learning based object detector~\cite{Girshick15ICCV,He17ICCV,Redmon18arXiv}.
However, the limited field of view of such cameras limits the detection to a narrow frustum.
Furthermore, the inaccurate depth measurements at far ranges make accurate person localization difficult in 3D space.
Instead, a 2D LiDAR provides accurate range measurements with a large field of view at high acquisition rates.
Thus, it is a promising sensor choice for detecting persons.

However, the limited information contained in the sparse range measurements from a 2D LiDAR is a key challenge towards reliable person detection.
Recent developments have shown that it is beneficial to combine several previous scans in order to detect objects~\cite{Beyer18RAL,Zhu19arXiv}.
In particular, Beyer~\etal~reported improved detection results by accumulating five previous scans, compared to their single scan baseline~\cite{Beyer18RAL}.
The downside, however, is the increased computational cost.
Due to the ego-motion of the LiDAR, as well as the motion of objects in the scene, scans recorded at different times are not perfectly aligned, and an expensive alignment operation has to be carried out in order to fuse scans for downstream processing.
In the case of~\cite{Beyer18RAL}, the alignment is done using the odometry information in addition to repetitive sampling on previous scans.
This alignment has linear computational cost with respect to the number of scans, and using five previous scans already makes the overall detection pipeline too expensive for real-time processing on mobile platforms. 

  

\begin{figure}
\centering
\begin{overpic}[scale=0.23,tics=10]{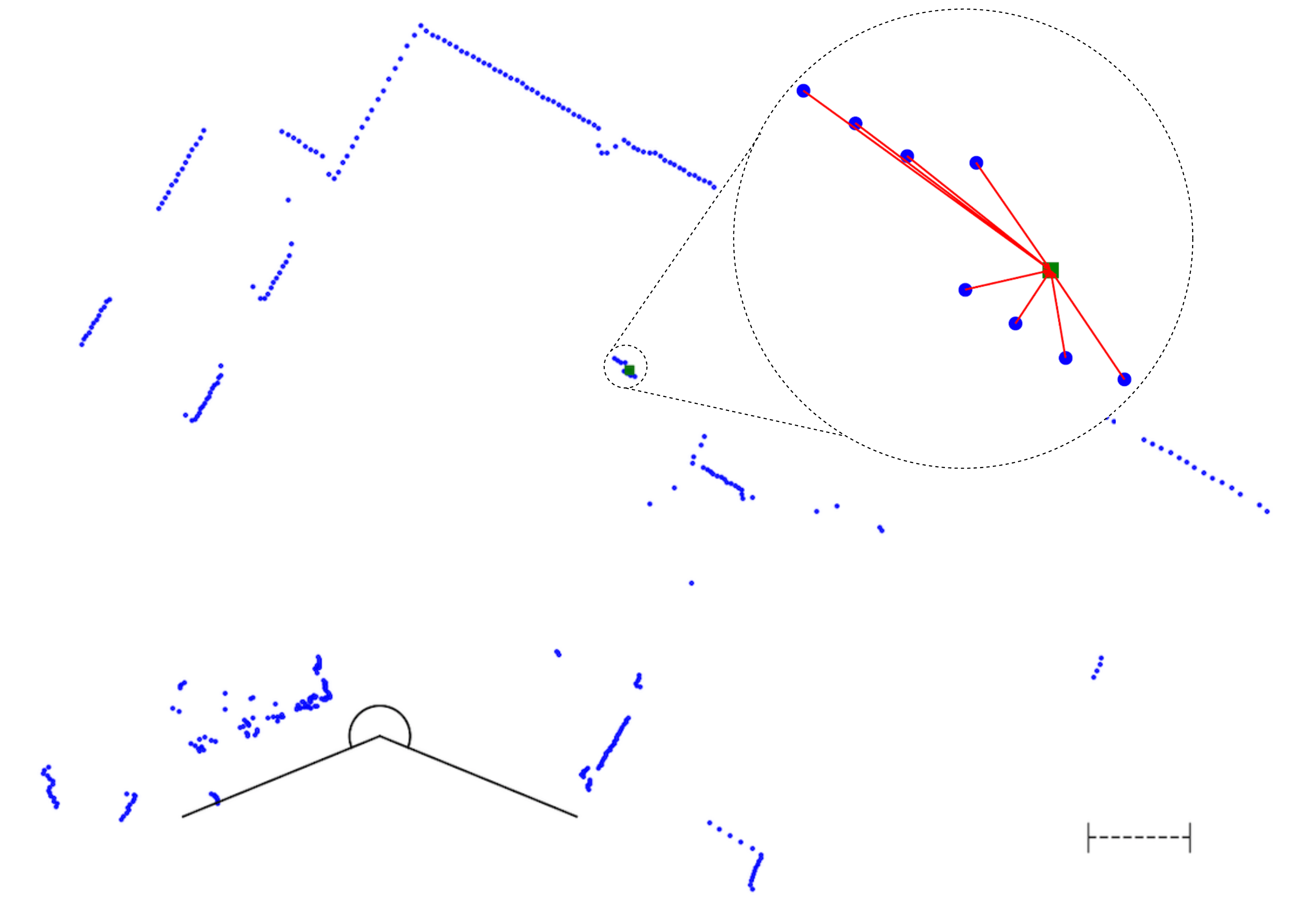}
\put(20.5,8){\footnotesize 2D LiDAR}
\put(85.1,8){\footnotesize 1\,m}
\end{overpic}
  
  \caption{The bird's-eye view of a 2D LiDAR scan (blue dots) with a person in the scene (green square). For each point \ourmethod~outputs a classification label and an offset to the center of the person (red lines), which are then grouped into detections using a post-processing step.}
  \label{fig:teaser}
\end{figure}

In principle, aligning and fusing previous scans follows a backward looking paradigm for aggregating temporal information.
In contrast, a forward looking paradigm simply keeps a representation based on current measurements and recurrently updates this representation when a new measurement becomes available.
This representation, which incorporates all previous measurements, is then used for downstream processing.
An example of such a forward looking paradigm is found in the field of video object detection, where researchers have used memory modules which recurrently take input features at each frame of a video sequence and output a refined prediction at the current frame~\cite{Tripathi16arXiv,Lu17ICCV,Xiao17ECCV}.

In this paper, we propose a person detection network which aggregates temporal information following a forward looking paradigm.
Our method uses the existing architecture of the DROW detector~\cite{Beyer18RAL}, which takes as input a 2D scan and predicts for each point a classification label and an offset vector pointing towards the center of the person~(Fig.~\ref{fig:teaser}).
We augment DROW with an auto-regressive model to aggregate the intermediate features from the backbone network from each scan, allowing our method to use information from all past measurements.
Instead of explicitly aligning the intermediate features from different scans, we use a spatial attention mechanism to associate features from neighboring locations based on their similarity, significantly reducing the computational cost.
We call our method Distance Robust SPatial Attention and Auto-regressive Model (DR-SPAAM).
Evaluated on the original DROW dataset, DR-SPAAM outperforms the previous approaches, while running at 87.2~FPS on a laptop with a dedicated GPU, or at 22.6~FPS on an NVIDIA~Jetson~AGX.
The high frame rate of DR-SPAAM makes it suitable for many robotic applications.

In summary, we make the following key contributions:
\begin{itemize}
    \item We propose a spatial attention and auto-regressive model that fuses information from previous LiDAR scans without the need of explicit alignment operation.
    \item We propose \ourmethod, a fast 2D LiDAR based person detector using the spatial attention and auto-regressive model. Our proposed method outperforms the previous state-of-the-art in 2D range data based person detection both in speed and detection performance.
    \item We release our implementation in PyTorch, including a ROS node with pre-trained models, to facilitate easy deployment in robotic projects\footnote{https://github.com/VisualComputingInstitute/DR-SPAAM-Detector}.
\end{itemize}

\section{RELATED WORKS}
\label{sec:related_works}

\subsection{Person Detection from 2D Range Data}
Person detection from 2D range data has a long standing history in the robotics community.
While early methods are mainly based on heuristics to find specific shapes in range data, the most common paradigm is to segment a scan into connected segments, compute a set of hand-crafted features for each of these, and finally classify them to create detections~\cite{Arras07ICRA,Pantofaru10ROS,Leigh15ICRA}.
Common methods can be divided into approaches that detect, and then track, individual legs in order to detect persons \cite{Pantofaru10ROS,Leigh15ICRA}, or approaches that directly aim to include both legs in one segment \cite{Arras07ICRA}.
Optimally, one would learn the representation of persons directly from data to avoid making such hard design choices. 
The DROW detector \cite{Beyer16RAL} was the first deep learning based walking aid detector working on 2D range data and was later extended to additionally detect persons \cite{Beyer18RAL}.
One key aspect that improved the person detection results was the integration of temporal information.
However, this significantly increases the runtime, making it infeasible for online use.
We propose a new person detector that combines the existing architecture of the DROW detector with a forward looking temporal integration module, outperforming the original DROW version both in speed and in detection quality.



Others have used deep learning based methods on 2D range data.
Ondr{\'u}{\v{s}}ka \etal \cite{Ondruska16RSSWDL} create an occupancy grid from a stationary 2D LiDAR and predict class labels and future grid configurations based on RNNs for every grid cell.
In a later version they extend this approach to work with moving LiDARs~\cite{Dequaire18IJRR}.
However, in both cases they do not create object detections or tracks and as such cannot be compared to a person detector.

\subsection{Object Detection in 3D Point Clouds}
Many works have focused on detecting objects in point clouds obtained from a 3D LiDAR, since such a task plays an important role in autonomous driving applications. 
Point clouds, as a data representation, inherently lack the definition of structure and neighborhoods, and thus prohibit the use of popular CNN architectures. 
To solve this problem, earlier works have leveraged image-based object detection methods, either by projecting the point cloud onto an image plane, or by popping up 2D detections made on the RGB image with known extrinsic calibration~\cite{Chen16CVPR,Qi18CVPR,Ku17IROS,Lang19CVPR}.
The runtime and accuracy of these methods, however, are bottlenecked by the employed 2D object detector.
Later developments in the field, including current state-of-the-art methods, utilize the full point cloud without projection. 
Methods like VoxelNet\,\cite{Zhou17CVPR}, or SECOND\,\cite{Yan18Sensors} run (sparse) convolutions \cite{Graham18CVPR, Choy19CVPR} on structured 3D voxel grids converted from point clouds, while other methods like PointRCNN\,\cite{Shi19CVPR} directly process unstructured point clouds, using PointNet\,\cite{Qi17CVPR} inspired backbones \cite{Qi17NIPS, Engelmann20ICRA, Wang18TOG}.
Yet others combine both design schemes into two-stage detectors \cite{Chen19arXiv,Yang19ICCV,Shi19arXiv}.

A particularly interesting work is the VoteNet proposed by Qi~\etal\cite{Qi19ICCV}.
Given a 3D point cloud, the first stage of the VoteNet regresses for each point an offset vector towards the object center (a vote), similar to the DROW detector, but instead of using a post-processing step, it utilizes another sub-network to group the per-point predictions into bounding box proposals.
Thus, the whole network is end-to-end trainable.
Similarly, 3D-MPA~\cite{Engelmann20CVPR} uses voting for instance proposals which are grouped into point-level instance masks.
This differentiable vote aggregation could also be an interesting approach for 2D LiDARs.
However, our focus lies on the temporal integration of a sequence of scans, and these two approaches are orthogonal.


Although from a hardware point of view, a 2D LiDAR does share similar working principles with its 3D counterpart, the methods operating on 3D point clouds cannot be naively applied to range data obtained from a 2D LiDAR.
To the best of our knowledge, no work exists that directly applies existing 3D methods to 2D range data, and given the sparsity of information in 2D range scans, it remains to be seen if a naive adaptation is possible at all.

\subsection{Video Object Detection}
Video object detection is a special object detection task, where the input is a video sequence showing the same objects in multiple frames.
It is thus possible to utilize information from one frame to aid detection in another.
This temporal propagation of information is important, since objects can undergo large appearance changes caused by fast motion, occlusion, or change of camera angle.
Earlier approaches~\cite{Kang16CVPR,Han16arXiv,Feichtenhofer17ICCV} detect objects in each frame independently and apply sequence-level post-processing on the obtained bounding boxes.
Such approaches cannot be optimized in an end-to-end fashion.
Later approaches focus on directly aggregating features across frames, either explicitly aligning features using optical flow~\cite{Zhu17CVPR,Zhu17ICCV,Wang18ECCV}, or using a memory network to aggregate features~\cite{Tripathi16arXiv,Lu17ICCV,Xiao17ECCV}. 

Similar to these approaches, our method uses a network to aggregate features across consecutive scans.
Instead of using a designated memory module, we use an auto-regressive model, which propagates information from the previous scan to the next with an exponentially decaying weight.
In video object detection, the key is to be able to aggregate long term features, since neighboring frames often have similar appearance and introduce little new information.
Thus, a memory network is often used.
Instead, we focus on aggregating short term features from consecutive scans, with the goal of enriching the information available for detection.
Hence, an auto-regressive model is more suitable compared to a more complex memory module.
Furthermore, we propose to use a similarity based spatial attention model~\cite{Wu19ICCV, Vaswani17NIPS} to first fuse nearby spatial information before the temporal aggregation.


\section{METHOD}
\label{sec:methodology}

In this section we describe in detail our proposed method for person detection.
We first describe our baseline architecture -- the DROW detector~\cite{Beyer18RAL}, followed by a discussion on different paradigms of aggregating temporal information.
In the end, we introduce our proposed \ourmethod~detector, which in contrast to~\cite{Beyer18RAL}, uses a forward looking approach to aggregate information over time.

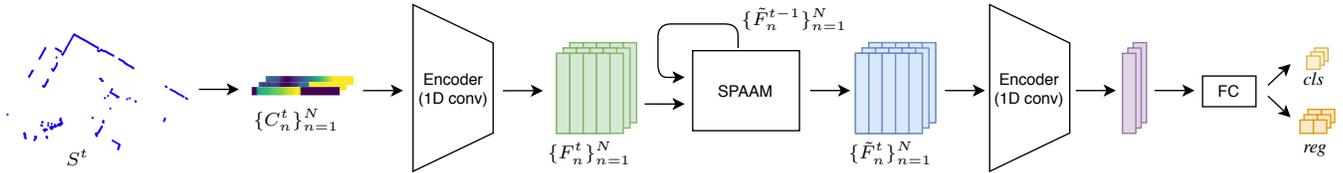
\begin{figure*}[t]
  \centering
  \begin{overpic}[width=\textwidth,tics=10]{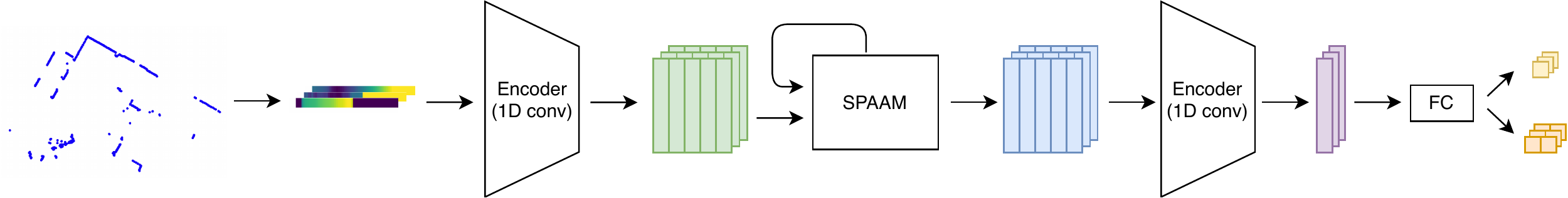}
    \put(5,0.5){\scriptsize $S^t$}
    \put(19,3.5){\scriptsize $\{C_n^t\}_{n=1}^N$}
    \put(41,1){\scriptsize $\{F_n^t\}_{n=1}^N$}
    \put(55.5,11){\scriptsize $\{\tilde{F}_n^{t-1}\}_{n=1}^N$}
    \put(63.5,1){\scriptsize $\{\tilde{F}_n^t\}_{n=1}^N$}
    \put(97.5,1.5){\scriptsize \textit{reg}}
    \put(97.5,6.5){\scriptsize \textit{cls}}
  \end{overpic}

  \caption{Overview of the \ourmethod~architecture.
  From the current LiDAR scan we create cutouts $C_n^t$ for every point, from which the network computers intermediate representations $F_n^t$.
  Using our SPAAM module, we aggregate temporal information from previous scans (\cf Fig.~\ref{fig:attentation} and Sec.~\ref{sec:detector_details}).
  Based on the merged representations $\tilde{F}_n^t$ the network outputs a classification label and predicts a relative object center for each point.
  }
  \label{fig:dr_spaam_arch}
\end{figure*}

\subsection{DROW Detector}
The DROW detector~\cite{Beyer16RAL,Beyer18RAL} was the first deep learning-based approach that detects persons from 2D range data.
It consists of three stages.
First, in order to normalize the appearance across different depths, the raw scans are preprocessed into small per-point windows, which are referred to as \textit{cutouts}.
These cutouts are then separately classified by a network as either object or background, and a possible object center is regressed for every cutout.
Finally, all regressed object centers (referred to as \textit{votes}) are collected and aggregated to a final set of detections.


During preprocessing, given a scan $S^t\in\mathbb{R}_{>0}^N$ composed of $N$ points\footnote{
A 2D LiDAR scan is composed of range measurements at different angles. For simplicity, we refer to a range measurement $s$ as a point, omitting the angular component.} 
$\{s_n^t\}_{n=1}^N$ at time $t$, $N$ cutouts $\{C_n^{t}\}_{n=1}^N$ are generated, each corresponding to a fixed-size window in Euclidean space around the LiDAR point $s_n^t$. 
This is done by computing an angular opening $a_n^t$ for each point $s_n^t$ using
\begin{equation}
\label{eqn:angle}
    a_n^t = 2 \cdot \arctan{\frac{0.5 \cdot \widebar{W}}{s_n^t}},
\end{equation}
where $\widebar{W}$ is a hyperparameter specifying the cutout width.
The points within this angular neighborhood are then re-sampled to a fixed number of $M$ points and  
centered by subtracting the distance $s_n^t$ of the central point.
Background and foreground points 
outside a depth range of $\pm d$
are clipped away and replaced with a constant value, based on the threshold $d$, and finally all values in the cutout $C_n^t$ are normalized to $[-1, 1]$.
The now normalized cutouts $\{C_n^{t}\}_{n=1}^N$ are passed through a network for classification and regression.
During postprocessing, votes are accumulated in a voting grid and a non-maximum suppression step is applied to obtain a set of detections, which are further refined by aggregating the class distributions of votes belonging to a detection.
The results of the DROW detector show the benefit of the cutout preprocessing.
In particular, it alleviates problems caused by unequal sampling densities at different distances (LiDAR points are sparser at far range), and also allows the DROW detector to work with LiDARs with different angular resolutions without requiring re-training.
Furthermore, the clipping operation removes the background information, allowing the network to focus on the neighboring points that are in the same distance region.
We refer the reader to~\cite{Beyer16RAL} for more details.

\subsection{Temporal Information Aggregation}

Since the measurements obtained from LiDAR sensors are low in information content (especially at far range), some detectors aggregate measurements made at different times to obtain a richer representation of the space, and it has been observed that this temporal aggregation improves the performance of downstream tasks~\cite{Beyer18RAL,Zhu19arXiv}.
Many common techniques for accumulating the temporal information follow the so-called backward looking paradigm, where measurements within the past few steps are combined together. 
Spatial misalignment often exists between these measurements, due to the sensor ego-motion or dynamic objects, and this misalignment has to be corrected based on odometry or point cloud registration.

Similarly, the second version of the DROW detector also accumulates temporal information by looking backward~\cite{Beyer18RAL}.
It computes cutouts on the past $T$ scans $\{C_n^{t-T},\cdots,C_n^t\}$ and fuses features $\{F_n^{t-T},\cdots,F_n^t\}$, obtained from an intermediate stage of the network, by a simple summation.
The fused features are then fed into the later stage of the network for classification and regression.
Due to the ego-motion of the sensor, two range measurements $s_n^t$ and $s_n^{t-1}$ made at the same angular index $n$ will not correspond to a single aligned point in the world, and the DROW detector uses robot odometry to correct the misalignment before fusing the features $F_n^t$ and $F_n^{t-1}$.
However, odometry alone is not sufficient to compensate for the misalignment caused by dynamic objects in the scene.
In the case of persons, this is especially critical, since the LiDAR ray at the same $n^{th}$ angular index could hit the leg of a person at time $t-1$, while passing between the legs and hitting a distant background structure at time $t$, resulting in significantly different features.
Thus Beyer~\etal propose to fix the sampling location at which the cutout is centered to the location of the current point $s^t_n$. However, this means the cutouts of previous scans need to be recomputed at each time step.
With the alignment using odometry and fixed location sampling, the DROW detector combines five scans from the past, resulting in improved detection accuracy compared to using only a single scan.

The performance gain from such a backward looking approach comes at the cost of increased computation time.
The misalignment between the current measurement and each previous measurement has to be corrected, resulting in a linear increase in computation time with respect to the number of frames in the aggregation window.
For the DROW detector, using only five scans already makes the overall detection pipeline too expensive for real-time applications on mobile platforms.

An alternative approach to aggregate temporal information is to follow a forward looking paradigm.
Instead of explicitly aligning and combining multiple previous measurements, a forward looking approach simply keeps a representation based on the current measurements and recurrently updates the representation for each new measurement.
Ideally, the update step only incurs a small computational overhead.
As a result, a forward looking approach aggregates information from the past without the unfavorable runtime scaling behavior with respect to the size of the temporal window.

\subsection{\ourmethod~Detector}\label{sec:detector_details}
\begin{figure}
  \centering
  \begin{overpic}[scale=0.65,tics=10]{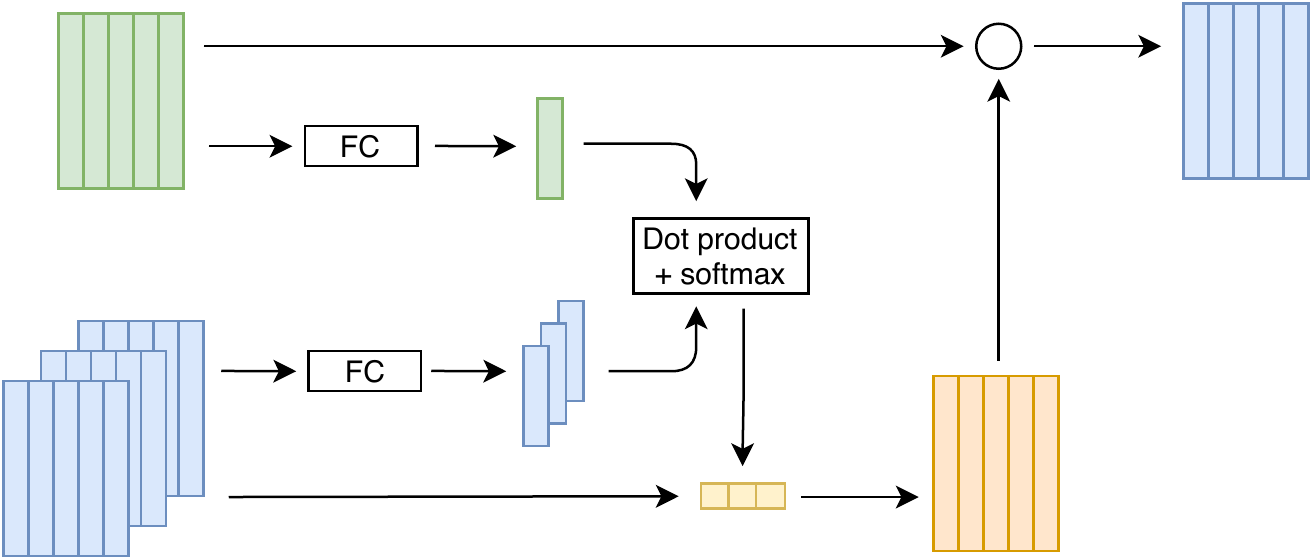}
    \put(7,25){\scriptsize $F_n^t$}
    \put(11,-3){\scriptsize $\{\tilde{F}_{n-w}^{t-1}\cdots \tilde{F}_{n+w}^{t-1}\}$}
    \put(58, -3){\scriptsize $\sum_{j=n-w}^{n+w}\mathit{softmax}(\tilde{\Omega}_{n})_{j}\tilde{F}_j^{t-1}$}
    \put(52.2,1.6){\scriptsize \textit{weights}}
    \put(93,25){\scriptsize $\tilde{F}_n^{t}$}
    \put(78,33){\scriptsize $1-\alpha$}
    \put(68.5,37){\scriptsize $\alpha$}
    \put(74.5,37.7){\Large +}
  \end{overpic}
  \vspace*{0.1cm}
  
    \caption{The SPAAM module associates features from previous scans using similarity-based spatial attention. 
    The similarity-weighted features are then combined with the current features in an auto-regressive fashion.}
  \label{fig:attentation}
\end{figure}
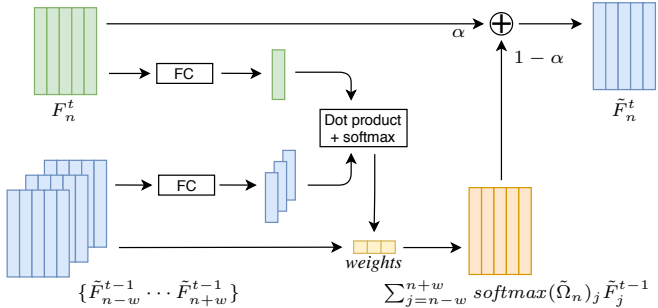

We propose the Distance Robust Spatial-Attention and Auto-regressive Model (\ourmethod), which follows a forward looking paradigm to aggregate temporal information.
Instead of computing spatially aligned cutouts on the past scans, we use a similarity-based spatial attention module~\cite{Vaswani17NIPS}, which allows the network to learn to associate misaligned features from 
a spatial neighborhood.
Additionally, an auto-regressive model is used to update the representation, aggregating information forward through time.
Our proposed detector outperforms the DROW detector, while being approximately four times faster.
Diagrams of \ourmethod~and of the proposed spatial attention and auto-regressive model are shown in Fig.~\ref{fig:dr_spaam_arch}~and~\ref{fig:attentation} respectively.

Due to the misalignment, features $F_n^t$ and $F_n^{t-1}$ computed at two time steps cannot be naively combined.
Instead of explicitly modeling the alignment as in~\cite{Beyer18RAL}, we propose to let the network learn to associate features using a similarity-based attention mechanism.
For a point $s_n^t$, we look at its spatial neighbors $\{s_{n-w}^{t-1},\cdots,s_{n+w}^{t-1}\}$ at previous time $t-1$ and compute a pairwise similarity
\begin{equation}
\label{eqn:similarity}
    \Omega_{nj}=\psi(F_j^{t-1})^T\cdot\psi(F_n^t)
\end{equation} 
between the features extracted at each previous point $s_j^{t-1}$ and those from the current point $s_n^t$.
Here, $w$ is a parameter defining the size of the neighborhood, $F_n^t$ is the intermediate feature of $s_n^t$, and $\psi$ is a generic mapping function, realized by a neural network, that maps the feature to an embedding space.
We then use a softmax function to convert the similarity into weighting factors and produce fused features $\tilde{F}_n^{t-1}$ from the previous frame: 
\begin{equation}
\label{eqn:attention}
    \tilde{F}_n^{t-1}=\sum_{j=n-w}^{n+w}\mathit{softmax} (\Omega_{n})_{j}F_j^{t-1}.
\end{equation}
This model gives more weight to the points with a higher similarity score, which are more likely to contain information from regions near $s_n^t$, while suppressing features from other irrelevant points.
The fused feature $\tilde{F}_n^{t-1}$ from the previous frame can then be combined with the current feature $F_n^t$ and be used for further processing. 

This model, however, only combines information from two consecutive scans.
In order to aggregate information from previous scans further back in the past, we propose to combine Eqn.~\ref{eqn:attention} with an auto-regressive approach.
We treat the fused features $\tilde{F}_n^{t-1}$ from time $t-1$ as a template, and when the new features $F_n^t$ at time $t$ become available, we compute an updated template:
\begin{equation}
    \tilde{F}_n^{t} = \alpha F_n^{t} + (1-\alpha) \sum_{j=n-w}^{n+w}\mathit{softmax}(\tilde{\Omega}_{n})_{j}\tilde{F}_j^{t-1},
\end{equation}
where $\alpha \in [0, 1)$ is a parameter that controls the update rate.
Here the first term is our update to the stored template, and the second term summarizes the information from the past.
Notice that unlike in Eqn.~\ref{eqn:similarity}, the term $\tilde{\Omega}_{nj}$ here denotes the similarity between the current feature $F_n^t$ and the neighboring features $\{\tilde{F}_{n-w}^{t-1},\cdots,\tilde{F}_{n+w}^{t-1}\}$ from the previous templates, rather than from the previous scan, \ie
\begin{equation}
    \label{eqn:spaam_similarity}
    \tilde{\Omega}_{nj}=\psi(\tilde{F}_j^{t-1})^T\cdot\psi(F_n^t).
\end{equation}
The updated template $\tilde{F}_n^t$ is then passed to the later stage of the network for the final classification and offset regression.

Compared to the original DROW detector, our \ourmethod~detector has a significantly lower computational complexity, requiring neither robot odometry nor the cutout recomputation on previous scans.
The auto-regressive model also allows our detector to keep only a single template per angular index, without having to store multiple past scans, while being able to accumulate information within a larger temporal window. 

\section{EVALUATION}
\label{sec:evaluation}

We evaluate our methods using the DROW dataset~\cite{Beyer16RAL, Beyer18RAL}, which is recorded in an indoor rehabilitation facility using a SICK S300 scanner.
The dataset contains 24,012 annotated scans, split into \textit{train} (17,665), \textit{validation} (3,919), and \textit{test} (2,428) sets.
The annotation includes the locations of three classes of objects: \textit{wheelchair}, \textit{walker}, and \textit{person}.
In this work we specifically focus on detecting persons and ignore the annotations of the other two categories, albeit our method should be general enough to handle other classes with adjusted hyper-parameters.

Following the standard in the object detection community, we use average precision (AP) at different association distances as our main evaluation metric.
AP$_d$ means that a detection is considered as positive if there exists an unmatched ground truth that is within $d$\,m radius of the predicted location.
Notice that~\cite{Beyer16RAL, Beyer18RAL} reported the area under the precision-recall curve (AUC), which is equivalent to average precision by definition. Additionally, we report the peak-F1 score (using 0.5\,m association distance), which is the maximum harmonic mean of the different precision and recall values, as well as the equal error rate (EER), the value at which precision and recall are equal.

All our models are trained on the \textit{train} set with a batch size of 8 scans for 40 epochs.
For our \ourmethod~detector, we load 10 frames back into the past during training, since scans that are further back in time are less relevant due to the exponential decay.
We use an Adam optimizer, with initial learning rate $10^{-3}$ with an exponential decay (after each iteration) to $10^{-6}$ during the complete training.
For the classification we use the binary cross entropy loss and for the regression we use the $L1$-norm of the regression error.
To convert the network output into detections we use the same postprocessing scheme introduced in~\cite{Beyer18RAL}.
We use \textit{Hyperopt}~\cite{Bergstra13ICML} to optimize the hyper-parameters of the voting step by maximizing AP$_{0.5}$ on the validation set for each model individually.
During evaluation, similar as during training, we provide a temporal context of 10 past frames for each test scan. However, our approach readily generalizes to run on complete sequences.

\subsection{Quantitative Results}

\begin{table}[t]
\centering
\caption{Detection accuracy on the test set with 0.5$m$ association threshold. Note that \ourmethod~and our re-trained baseline DROW do not use odometry information.}
\begin{tabular}{ p{3.6cm} c c c c }
\toprule
Method && AP$_{0.5}$ & peak-F1 & EER \\ 
\midrule
ROS leg detector~\cite{Pantofaru10ROS} && 23.2 & 41.7 & 41.0\\
Arras~(re-trained)~\cite{Arras07ICRA} && 47.6 & 50.3 & 50.1\\
Leigh~(re-trained)~\cite{Leigh15ICRA} && 57.2 & 64.3 & 62.3\\
DROW ($T=1$) in~\cite{Beyer18RAL} && 59.4 & 61.5 & 61.4\\
DROW ($T=5$) in~\cite{Beyer18RAL} && 67.0 & 65.9 & 64.9\\
DROW ($T=5$, + odom.) in~\cite{Beyer18RAL} && 68.1 & 68.1 & \textbf{67.2}\\
\arrayrulecolor{lightgray}\midrule[0.25pt]\arrayrulecolor{black}
DROW ($T=1$) baseline && 66.6 & 66.1 & 65.2\\
DROW ($T=5$) baseline && 67.9 & 65.1 & 63.8\\
DR-AM (w/o spatial attention) && 66.3 & 65.2 & 64.0\\
DR-SPA (w/o auto-regression) && 68.0 & 67.0 & 66.1\\
\ourmethod && \textbf{70.3} & \textbf{68.5} & \textbf{67.2}\\
\bottomrule
\end{tabular}
\label{table:results}
\end{table}

We evaluate our proposed method using the test set and report the average precision, peak-F1 score, and equal error rate of our method at an association threshold of 0.5\,m in Table~\ref{table:results}.
As an additional baseline, we also report the performance of two re-trained DROW models, using a single scan and five scans, respectively.
Compared to the original implementation, our re-trained models use a smaller cutout window and more sampling points within each cutout, which we selected based on a better performance on the validation set (\cf Sec.~\ref{sec:abblation}).
In order to keep the comparison meaningful, the same cutout parameters are used for both the re-trained DROW baseline and \ourmethod, and no odometry information is used.
The original DROW score reported by Beyer~\etal on the person class in~\cite{Beyer18RAL}, as well as the score of the Leigh~\cite{Leigh15ICRA} and Arras detectors~\cite{Arras07ICRA}, re-trained on the DROW dataset, are also included in Table~\ref{table:results}.

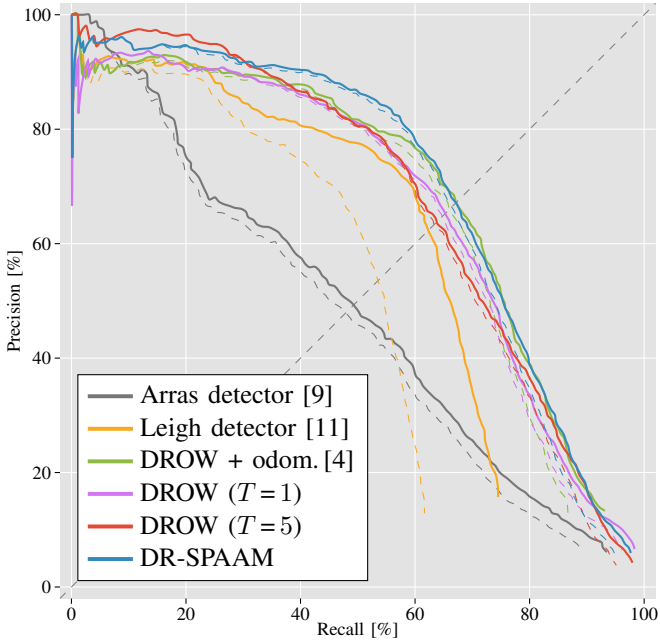
\begin{figure}[t]
    \centering
    \begin{tikzpicture}[]
    \begin{axis}[
    precrec, legend cell align={left}, legend style={fill=white!11!white, draw=black}, legend pos=south west]
    \draw [gray,dashed] (rel axis cs:0,0) -- (rel axis cs:1,1);
    \precrec{arras_retrained}{good_gray}{smooth, thick}\label{plot:arras_retrained}
    \precrecnl{arras_retrained_0.3}{good_gray}{smooth, dashed}\label{plot:arras_retrained_r}
    \addlegendentry{Arras detector~\cite{Arras07ICRA}}
    \precrec{leigh_retrained_hyperopted}{awesome_orange}{smooth, thick}\label{plot:leigh_retrained}
    \precrecnl{leigh_retrained_hyperopted_0.3}{awesome_orange}{smooth, dashed}\label{plot:leigh_retrained_r}
    \addlegendentry{Leigh detector~\cite{Leigh15ICRA}}
    \precrec{WNet3xLF2p-T5-odom=rot_wp}{wp_c}{smooth, thick}\label{plot:drow3xlf}
    \precrecnl{WNet3xLF2p-T5-odom=rot_wp_0.3}{wp_c}{smooth, dashed}\label{plot:drow3xlf_r}
    \addlegendentry{DROW + odom.\,\cite{Beyer18RAL} }
    \precrec{drow_bl_T1_0.5}{wa_c}{smooth, thick}\label{plot:drow_single}
    \precrecnl{drow_bl_T1_0.3}{wa_c}{smooth, dashed}\label{plot:drow_single_r}
    \addlegendentry{DROW~($T$\,=\,$1$)}
    \precrec{drow_bl_T5_0.5}{wd_c}{smooth, thick}\label{plot:drow_multi}
    \precrecnl{drow_bl_T5_0.3}{wd_c}{smooth, dashed}\label{plot:drow_multi_r}
    \addlegendentry{DROW~($T$\,=\,$5$)}
    \precrec{new_spaam_0.5}{wc_c}{smooth, thick}\label{plot:ss}
    \precrecnl{new_spaam_0.3}{wc_c}{smooth, dashed}\label{plot:ss_r}
    \addlegendentry{\ourmethod}
    \end{axis}
    \end{tikzpicture}
    \caption{Precision-recall curves for several baselines and our \ourmethod~detector, evaluated with association distance 0.5\,m (solid) and 0.3\,m (dashed).}
    \label{fig:naive_pr}
\end{figure}

As the results show, \ourmethod{} achieves the highest AP$_{0.5}$ of $70.3\%$, which is $2.4\%$ above the baseline model and $2.2\%$ above the original DROW in~\cite{Beyer18RAL}.
Thus, \ourmethod{} establishes a new state-of-the-art, even though it does not use odometry information.
By comparing our re-trained DROW models with the original DROW, we can also observe the effectiveness of our proposed adjustment, especially in the single scan case.
Fig.~\ref{fig:naive_pr} shows the precision-recall curves of all different models.
Here we can see that \ourmethod{} outperforms all other setups, except for a small region in the high precision regime, where the DROW (T=5) baseline scores higher.

Table~\ref{table:results} also shows the results of an ablation study that highlights the contribution of the different components in our temporal aggregation module.
DR-AM corresponds to a network where the auto-regressive model
is updated with the new features directly, without using the weighted sum from the spatial attention mechanism.
It has a slightly worse performance compared to a single-scan DROW baseline, showing that it is not beneficial to naively combine misaligned features. 
DR-SPA, on the other hand, only combines the features from the current and the previous scan using spatial attention, without using the accumulated feature template from the auto-regressive model.
This two-scan approach already outperforms the five-scan DROW baseline, showing the benefit of using a learning-based approach to incorporate features from the previous measurements. 
The full model,~\ourmethod, outperforms the two-scan DR-SPA, showing the benefit of aggregating information over a larger temporal window.

\subsection{Inference Time}
\ourmethod~not only achieves better detection, it also has a significantly lower computational complexity, since it eliminates the need to perform an expensive recomputation of past cutouts.
We implement all networks in Python and PyTorch without using any inference time acceleration framework (\eg, TensorRT), and profile the run time of different components of the complete pipeline.
The timing results for two mobile platforms are reported in Table~\ref{table:inference_time}.
Here we use a laptop equipped with a mobile NVIDIA GeForce RTX 2080 Max-Q GPU and an Intel-i7 9750H CPU, as well as a Jetson AGX Xavier.
The table shows that \ourmethod~has a computation time similar to that of a single-scan DROW method.
This is expected, since at each time step only the current scan needs to be processed and the more expensive fusion of~\ourmethod~adds a small overhead.
However, even though the methods are similar in speed, the single-scan DROW method has a by far lower detection accuracy.
DROW with temporal integration, on the other hand, has a runtime that scales linearly with the number of scans due to the required cutout recomputation on all past scans, and already becomes significantly slower when using five scans.
Considering the Jetson AGX platform, a frame rate of 2.6\,FPS is too slow for a real-time application, whereas our 9.7\,FPS is still well within a usable range.
On the laptop, we can in fact run all models faster than real-time, given that the DROW dataset was recorded at a frame rate of roughly 13\,FPS.
However, \ourmethod~can run at significantly higher frame rates if needed, providing detections with a lower latency and consuming less power, which is a relevant aspect for mobile platforms.

The cutout generation and the voting are the expensive steps in the whole pipeline.
To further increase the frame rate, we resort to a faster implementation of the cutout generation and the resulting model, \ourmethod$\ast$, runs at 87.2\,FPS on a laptop with a dedicated GPU, or at 22.6\,FPS on a Jetson AGX, well-beyond the requirement for many real-time applications.
The new implementation increases the network performance thanks to the improved numerical precision.


\begin{table}[t]
\centering
\caption{Computation time (in milliseconds) and frame rate of different setups on two different mobile platforms.}
\setlength{\tabcolsep}{2.5pt}
\begin{tabularx}{\linewidth}{l c cYYYccYYY }
\toprule
& & \multicolumn{4}{c }{Laptop (RTX 2080)} && \multicolumn{4}{c}{Jetson AGX} \\
\cmidrule{3-6} \cmidrule{8-11}
 Method & AP$_{0.5}$    & cutout & net & vote & FPS && cutout & net & vote & FPS\\ 
\midrule
DROW ($T$\,=\,$1$) & 66.6         & ~7.0 & 1.4 & ~6.1 & 68.6 && ~63.3 & 4.8 & 29.3 & 10.4\\
DROW ($T$\,=\,$5$) & 67.9         & 34.3 & 1.5 & 19.2 & 18.2 && 306.3 & 5.1 & 78.1 & ~2.6\\
\ourmethod & \textbf{70.3}        & ~7.0 & 2.0 & ~7.7 & 59.8 && ~62.0 & 6.9 & 33.6 & ~9.7\\
\midrule
\ourmethod $\ast$ & \textbf{71.8} & ~1.1 & 1.9 & ~8.5 & 87.2 && ~~4.2 & 7.7 & 32.4 & 22.6\\
\bottomrule
\end{tabularx}
\label{table:inference_time}
\end{table}


\subsection{Hyperparameter Selection}\label{sec:abblation}


\PAR{The cutout} operation is parameterized by the width ($\widebar{W}$) and depth ($D$) of the window, as well as the number of points ($N$) used for re-sampling. 
In~\cite{Beyer18RAL} a larger (1.66\,m$\times$2.0\,m) sized window with 48 points was used in order to cope with the bigger walking-aid classes.
Since we are only concerned with detecting persons, we propose to use a smaller sized window that tightly fits the footprint of a person, thus reducing distracting information from the surroundings.
We conduct an experiment on the size of the cutout window on the validation set using a DROW network with a single scan.
The results are shown in Table~\ref{table:cutout}.
Based on these results, we set our cutout window to (1.0\,m$\times$1.0\,m) with 56 points and use these cutout parameters for all models we train.

One can observe that the scores on the validation set are significantly lower than those on the test set.
The original DROW dataset is created for detecting three classes: \textit{wheelchair}, \textit{walker}, and \textit{person}.
In this work, we have only kept the annotations for the \textit{person} class, and we observed that the validation set happens to contain more person annotations at farther distances.
Even though we use a distance robust preprocessing, at farther distances information becomes so sparse that the detection will always become less reliable, thus rendering the validation set more challenging.

\PAR{The Spatial-Attention and Auto-Regressive Model} are parameterized by the update rate $\alpha$ and the size of the search window $W$.
Although the meaning of these two parameters is intuitively clear, it is not a trivial task to select the proper combination.
We train multiple \ourmethod~networks with different combinations of $\alpha$ and $W$.
Based on the validation set results (Table~\ref{table:spatial_similarity}), we choose to use $W$\,=\,$11$ and $\alpha$\,=\,$0.5$ for our final model.
No clear pattern can be seen in these results, neither larger search windows nor specific update rates consistently work better.
A more thorough search through the parameter space could potentially result in models that perform even better.

\begin{table}[t]
\centering
\caption{Validation set scores of DROW ($T$\,=\,$1$) detectors with different cutout parameters.}
\begin{tabular}{c c c c c c c c }
\toprule
$\widebar{W}$&  $D$ & $N$ &~& AP$_{0.3}$ & AP$_{0.5}$ & peak-F1 & EER\\ 
\midrule
1.66 & 2.0 & 48 && 41.9 & 43.0 & 48.1 & 47.6\\
1.66 & 1.0 & 48 && 42.6 & 43.4 & 49.2 & 48.6\\
1.0\phantom{0} & 2.0 & 48 && 43.6 & 44.8 & 50.7 & 50.4\\
1.0\phantom{0} & 1.0 & 48 && 44.0 & 45.0 & 50.3 & 50.2\\
\arrayrulecolor{lightgray}\midrule[0.25pt]\arrayrulecolor{black}
1.0\phantom{0} & 1.0 & 32 && 42.0 & 43.0 & 49.1 & 48.8\\
1.0\phantom{0} & 1.0 & 40 && 43.1 & 44.1 & 50.0 & 49.6\\
1.0\phantom{0} & 1.0 & 48 && 44.0 & 45.0 & 50.3 & 50.2\\
1.0\phantom{0} & 1.0 & 56 && \textbf{45.1} & \textbf{46.3} & \textbf{50.9} & \textbf{50.8}\\
1.0\phantom{0} & 1.0 & 64 && 43.8 & 45.1 & 50.7 & 50.4\\
\bottomrule
\end{tabular}
\label{table:cutout}

\bigskip

\caption{Validation set scores of \ourmethod~with different window sizes and update rates.} 
\begin{tabular}{ c c c c c c c }
\toprule
$W$&  $\alpha$\ &~& AP$_{0.3}$ & AP$_{0.5}$ & peak-F1 & EER\\ 
\midrule
7 & 0.3           && 45.0 & 46.2 & 52.5 & 52.5\\
7 & 0.5           && 49.5 & 50.9 & 54.6 & 53.6\\
7 & 0.8           && 46.8 & 48.3 & 54.1 & 54.0\\
\arrayrulecolor{lightgray}\midrule[0.25pt]\arrayrulecolor{black}
11 & 0.3                     && 51.5 & 53.0 & 56.8 & 56.4\\
11 & 0.5 && \textbf{52.7} & \textbf{53.9} & \textbf{57.3} & \textbf{57.3}\\
11 & 0.8                     && 47.4 & 48.7 & 53.6 & 53.2\\
\arrayrulecolor{lightgray}\midrule[0.25pt]\arrayrulecolor{black}
15 & 0.3                     && 51.5 & 52.8 & 56.1 & 55.3\\
15 & 0.5                     && 50.7 & 52.1 & 55.0 & 54.7\\
15 & 0.8                     && 47.0 & 48.2 & 53.1 & 53.0\\
\bottomrule
\end{tabular}
\label{table:spatial_similarity}
\end{table}

\subsection{Sampling Rate}
Different LiDARs often have different sampling rates.
Since a detection network is likely to be deployed on different LiDAR sensors, its robustness against varying sensor specifications should be examined.
We take two networks, DROW~($T$\,=\,$5$) and \ourmethod, and evaluate them on the test set using temporally sub-sampled sequences, simulating different sampling rates.
Table~\ref{table:temporal_spatial_stride} reports the detection accuracies at different temporal strides (a stride of $n$ means keeping only every $n^{th}$ scan).

The evaluation results show that \ourmethod~is very robust against changing sampling rate.
Even at a five times lower scanning frequency~(roughly 2\,Hz), the AP$_{0.3}$ only reduces by $2.1\%$.
This result shows the benefit of a learned spatial attention module, which combines information based on appearance similarity without relying on a fixed temporal context window.
Hence, \ourmethod~can be deployed on LiDARs with a wide range of sampling rates, or operate with a reduced sampling rate if the computation capacity is limited. 
On the other hand, the DROW detector performance degrades rapidly with increased temporal stride.
Larger time differences between consecutive scans lead to greater motion induced misalignments, which in turn hurt the accuracy of the DROW detector.

\begin{table}[h]
\centering
\caption{Test set results with different temporal strides.} 
\setlength{\tabcolsep}{3pt}
\begin{tabularx}{\linewidth}{c  cYYYY cYYYY}
\toprule
&& \multicolumn{4}{c}{DROW ($T$\,=\,$5$)} && \multicolumn{4}{c}{\ourmethod} \\
\cmidrule{3-6} \cmidrule{8-11} 
Stride && AP$_{0.3}$ & AP$_{0.5}$ & p-F1 & EER && AP$_{0.3}$ & AP$_{0.5}$ & p-F1 & EER\\
\midrule
1 && \textbf{66.6} & \textbf{67.9} & \textbf{65.1} & \textbf{63.8} && 68.5 & 70.3 & 68.5 & 67.2\\
2 && 59.3 & 60.5 & 60.1 & 59.3 && 69.3 & 70.8 & \textbf{68.8} & \textbf{67.6}\\
3 && 54.3 & 55.8 & 56.8 & 56.7 && \textbf{69.4} & \textbf{70.9} & 68.1 & 66.5\\
4 && 53.6 & 55.1 & 56.0 & 55.7 && 67.7 & 69.1 & 66.4 & 64.9\\
5 && 51.5 & 53.4 & 54.6 & 54.3 && 66.4 & 67.7 & 65.5 & 64.5\\
\bottomrule
\end{tabularx}
\label{table:temporal_spatial_stride}
\end{table}

\subsection{Temporal Association}

\begin{figure}
\begin{overpic}[scale=0.09,tics=10]{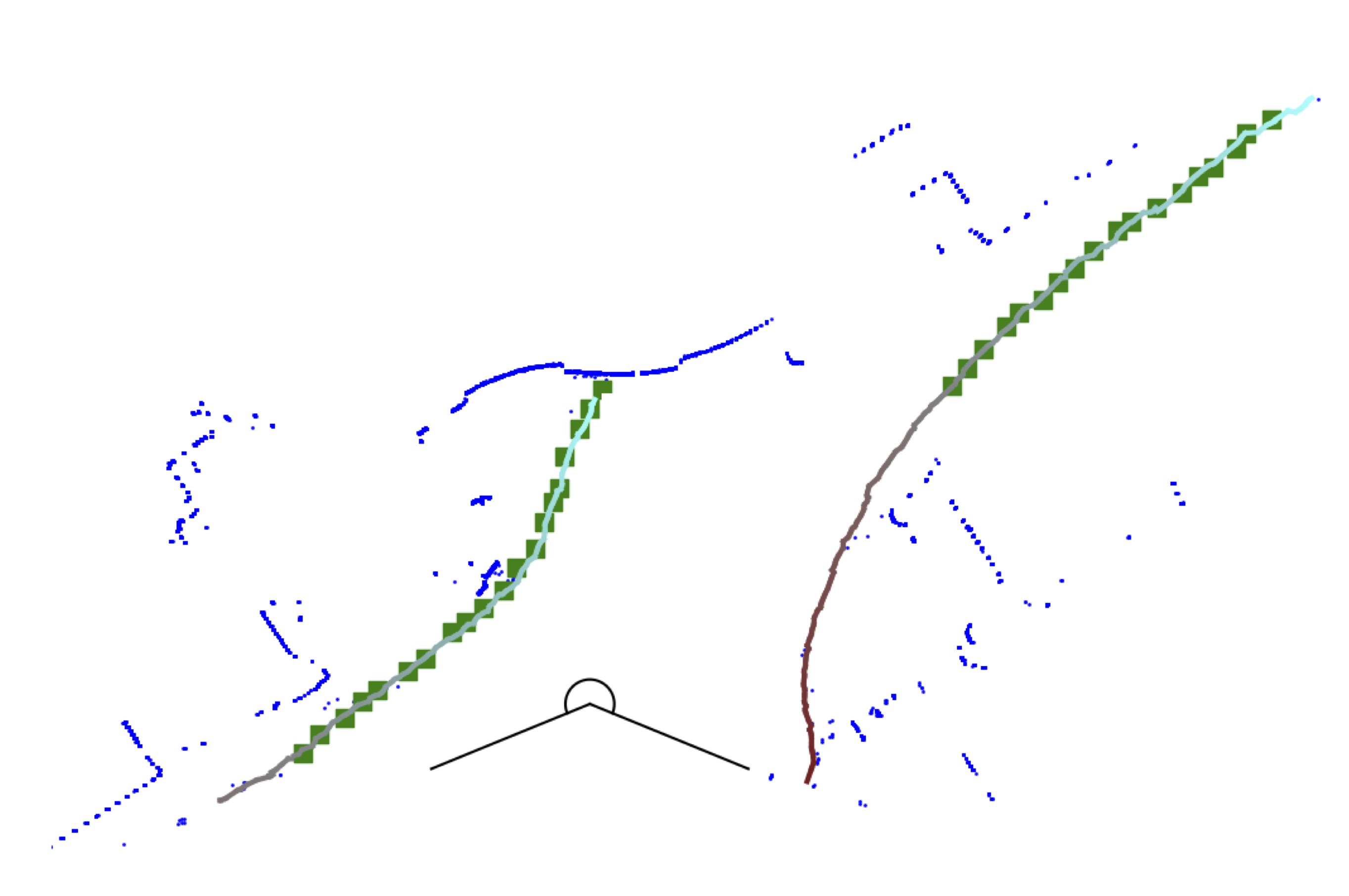}
\put(35.5,7){\footnotesize 2D LiDAR}
\end{overpic}
  \caption{Tracklets generated by \ourmethod~using 200 consecutive scans. 
  The blue points are the overlaid scans, and the green squares are the ground truth annotations.
  Notice that for clarity, we omit the points that have been classified as persons from plotting.
  The colored lines are the tracklets, and the coloring encodes the time of detections.
  The sequence is taken from the training set, since the validation set does not have any sequence recorded using a stationary LiDAR, which is needed for plotting overlaid scans.
  Nevertheless, most of the scans have not been annotated (shown as the missing annotations along the tracklets) and have not been exposed to the network during training.}
  \label{fig:tracking}
\end{figure}

During the temporal aggregation step, \ourmethod~computes the similarity between the aggregated template and the latest scan features~(Eqn.~\ref{eqn:spaam_similarity}).
This similarity can be further exploited for associating points across different scans.
Here we provide a preliminary example.
We take 200 consecutive frames (roughly spanning 15\,s) from a sequence and detect persons using~\ourmethod~for each individual frame.
For each detection $D_i^t$, we find its corresponding points in the previous scan, simply by selecting the ones with highest similarity score.
If these corresponding points were grouped into a detection $D_j^{t-1}$, and if the distance between the two detections are smaller than a threshold (0.5\,m), we group both detections into a tracklet.
Otherwise a new tracklet is started using $D_i^t$.
After 200 frames, we compute the confidence of each tracklet as the mean of all its detections.
In Fig.~\ref{fig:tracking} we plot all tracklets that have a  confidence greater than 0.35 and that are composed of at least five detections.
The trajectories of persons in the scene are clearly visible.
These associations can provide extra information for tracking algorithms. Their full potential is yet to be explored in future research.
The velocity and movement direction of the persons can also be derived from the associated detection pairs, and this information can be helpful for motion planning. 


\section{CONCLUSION}
\label{sec:conclusion}
We propose the \ourmethod~person detector that combines the distance robust detection scheme of the DROW detector with a powerful spatial attention and auto-regressive temporal integration model.
The spatial attention is able to associate misaligned features from different frames using their appearance similarity, while the auto-regressive model aggregates temporal information forward through time.
Compared to the previous state-of-the-art approaches, \ourmethod~achieves higher detection accuracy, while being significantly faster and able to run in real-time even on low-powered mobile platforms.
Experiments show that \ourmethod~generalizes well to LiDARs with different temporal sampling rates, and with our provided code and ROS node, we expect that our model will be useful for many robotic applications.

\textbf{Acknowledgements:}
We thank Francis Engelmann for his valuable feedback. 
This project was funded, in parts, by the EU H2020 project "CROWDBOT" (779942) and the BMBF project “FRAME” (16SV7830). 
Most experiments were performed on the RWTH Aachen University CLAIX 2018 GPU Cluster (rwth0485).












\balance

\bibliographystyle{ieeetr}
\bibliography{abbrev,mybib}

\end{document}